\pdfoutput=1
\documentclass[11pt]{article}
\usepackage{color,soul}
\usepackage{ACL2023}

\usepackage{times}
\usepackage{latexsym}
\usepackage[T1]{fontenc}

\usepackage[utf8]{inputenc}

\usepackage{microtype}

\usepackage{inconsolata}
\usepackage[utf8]{inputenc} 
\usepackage[T1]{fontenc}    
\usepackage{hyperref}       
\usepackage{url}            
\usepackage{booktabs}       
\usepackage{amsfonts}       
\usepackage{nicefrac}       
\usepackage{microtype}      
\usepackage{colortbl}
\usepackage{graphicx}
\usepackage[font=small]{caption}
\usepackage{subcaption}
\usepackage{wrapfig}

\usepackage{amsfonts}
\usepackage{amsmath}
\usepackage{amssymb}
\usepackage{amsthm}

\usepackage{algorithm}
\usepackage{algpseudocode}
\usepackage{booktabs}
\usepackage{multirow}
\usepackage{tabularx}
\usepackage{xcolor}

\usepackage{xspace}
\usepackage{enumitem}

\usepackage{scrextend}

\usepackage{makecell}
\usepackage{longtable}

\usepackage[normalem]{ulem}
\usepackage{pifont}

\newcommand{\cmark}{\ding{51}}
\newcommand{\xmark}{\ding{55}}

\newcommand{\tuple}[1]{\textsc{$\langle$#1$\rangle$}}

\title{BertNet: Harvesting Knowledge Graphs with Arbitrary Relations \\
from Pretrained Language Models}

\author{
Shibo Hao$^{1}$\thanks{~~Equal contribution. Code available at \url{https://github.com/tanyuqian/knowledge-harvest-from-lms}. Demo available at \url{https://lmnet.io}},~~
Bowen Tan$^{2*}$,~~
Kaiwen Tang$^{1*}$,~~
Bin Ni$^{1}$,~~
Xiyan Shao$^{1}$,~~\\
{\bf Hengzhe Zhang$^{1}$,~~
Eric P. Xing$^{2,3}$,~~
Zhiting Hu$^{1}$}\\
$^1$UC San Diego,~~ $^2$Carnegie Mellon University,\\ $^3$Mohamed bin Zayed University of Artificial Intelligence\\
{\small 
{\tt \{s5hao,zhh019\}@ucsd.edu}, \tt \{btan2\}@cs.cmu.edu 
}
}

\begin{document}
\maketitle
\begin{abstract}
It is crucial to automatically construct knowledge graphs (KGs) of diverse new relations to support knowledge discovery and broad applications. Previous KG construction methods, based on either crowdsourcing or text mining, are often limited to a small predefined set of relations due to manual cost or restrictions in text corpus. Recent research proposed to use pretrained language models (LMs) as implicit knowledge bases that accept knowledge queries with prompts. Yet, the implicit knowledge lacks many desirable properties of a full-scale symbolic KG, such as easy access, navigation, editing, and quality assurance. In this paper, we propose a new approach of harvesting massive KGs of \emph{arbitrary} relations from pretrained LMs. With minimal input of a relation definition (a prompt and a few shot of example entity pairs), the approach efficiently searches in the vast entity pair space to extract diverse accurate knowledge of the desired relation. We develop an effective search-and-rescore mechanism for improved efficiency and accuracy. We deploy the approach to harvest KGs of over 400 new relations from different LMs. Extensive human and automatic evaluations show our approach manages to extract diverse accurate knowledge, including tuples of complex relations (e.g., \texttt{"A is capable of but not good at B"}). The resulting KGs as a symbolic interpretation of the source LMs also reveal new insights into the LMs' knowledge capacities.

\end{abstract}

\section{Introduction}

Symbolic knowledge graphs (KGs) are a powerful tool for indexing rich knowledge about entities and their relationships, and are useful for information access \cite{googleKG}, decision making \cite{yang2021constructing,santos2022knowledge}, and improving machine learning in general \cite{Li2019KnowledgedrivenER,Wang2019MultiTaskFL,tan2020summarizing,Xiong2017ExplicitSR}.


\begin{figure}
    \centering
    \includegraphics[width=0.5\textwidth]{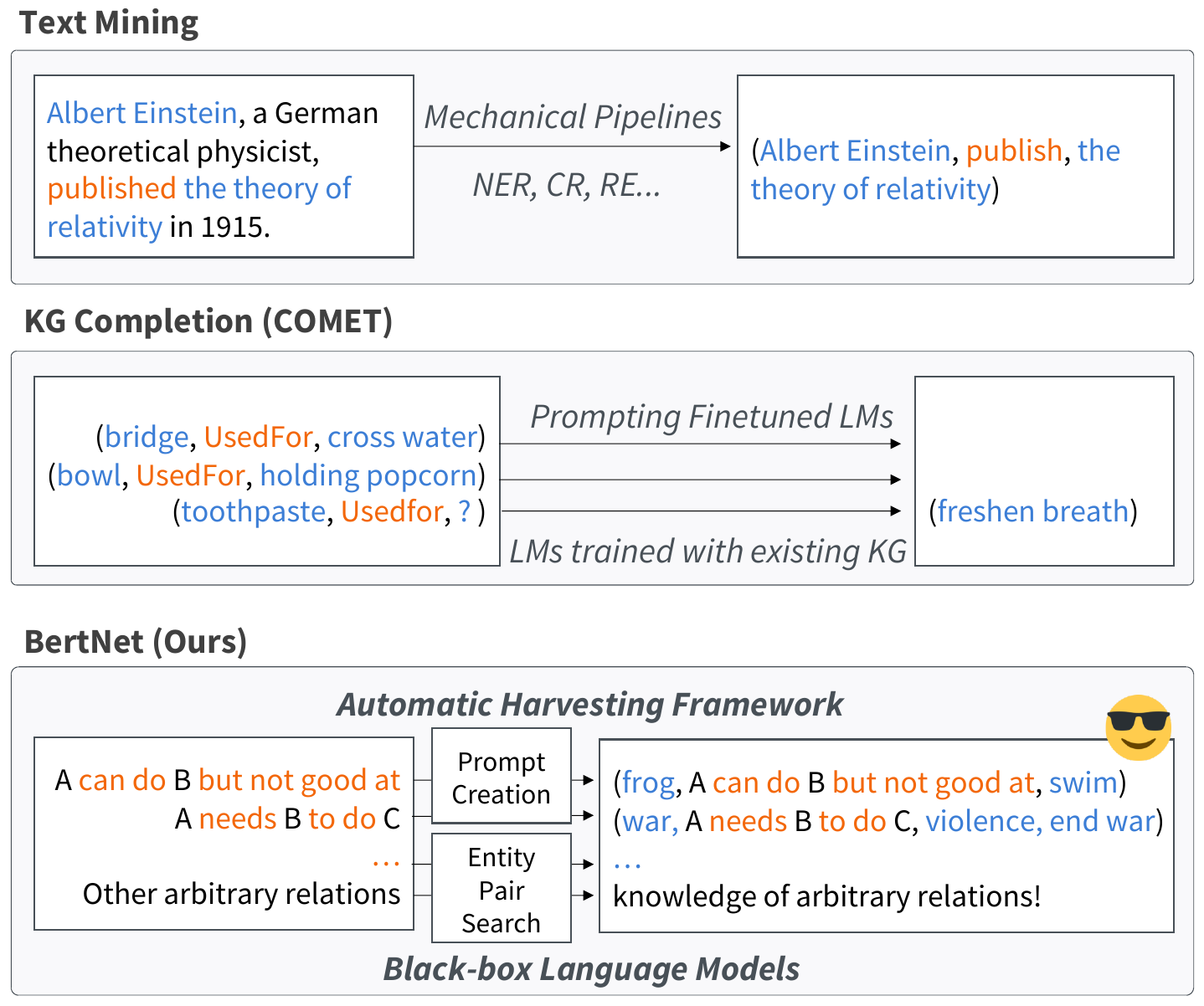}
    \vspace{-15pt}
    \caption{
    Different example paradigms of harvesting knowledge. \textit{Text mining} extracts knowledge of relations explicitly mentioned in the text. \textit{KG completion} produces tail entities to complete knowledge of preexisting relations. Our method is capable of harvesting knowledge of arbitrary new relations from LMs.
    }
    \label{fig:paradigms}
    \vspace{-5pt}
\end{figure}

\begin{table*}
\centering
\small
\begin{tabular}{@{}rllc@{}}
\toprule
Method           & Module(s)            & Outcome     & Arbitrary relation \\ \midrule
Text mining \cite{Zhang2020TransOMCSFL, nguyen2021refined}      & NER, CR, RE, etc.\footnotemark & KG          & \xmark                  \\ \midrule
LAMA \cite{Petroni2019LanguageMA}, LPAQA \cite{Jiang2020HowCW} & LMs                            & tail entity & \cmark                 \\
COMET \cite{Bosselut2019COMETCT}            & Finetuned GPT-2                  & tail entity & \xmark                  \\
Symbolic Knowledge Distillation \cite{west-etal-2022-symbolic}      & GPT-3                         & KG          & \cmark \footnotemark               \\ \midrule
BertNet (ours)   & LMs                            & KG          & \cmark                 \\ \bottomrule
\end{tabular}
\caption{Categorization of works on automatic knowledge extraction. Compared to other categories of approaches, our method extracts full \emph{explicit} KGs of \emph{arbitrary new relations} from \emph{any} LMs.
}
\label{tab:summary}
\vspace{-5pt}
\end{table*}

It has been a long-term desire to construct KGs of diverse \emph{relations} to comprehensively characterize the structures between entities. The traditional crowdsourcing-based approach \cite{Speer2017ConceptNet5A, Fellbaum2000WordNetA, Sap2019ATOMICAA} tends to cover only a restricted relation set, such as ConceptNet \cite{Speer2017ConceptNet5A} that contains a small set of 34 relations. The popular method based on text mining \cite{luan2019general, zhong2020frustratingly, Wang2021COVID19LK} has a similar limitation, as the text understanding models can often recognize only a predefined set of relations included in training data. Some open-schema text mining approaches (e.g., based on syntactic patterns) exist \cite{tandon2014webchild, romero2019commonsense, zhang2020aser, nguyen2021refined}, yet the extracted relations are limited to those explicitly stated in the text, missing all others that are not mentioned or do not have exact match with the text in the corpus. Similarly, KG completion approaches \cite{bordes2013translating, Bosselut2019COMETCT, Yao2019KGBERTBF} is restricted to the preexisting relations (Figure~\ref{fig:paradigms}).

On the other hand, large language models (LMs) pretrained on massive text corpus, such as \textsc{Bert} \cite{Devlin2019BERTPO} and \textsc{GPT-3} \cite{Brown2020LanguageMA}, have been found to encode a significant amount of knowledge implicitly in their parameters. Recent research attempted to use LMs as flexible knowledge bases by querying the LMs with arbitrary prompts (e.g., \texttt{"Obama was born in \uline{~~~}"} for the answer \texttt{"Hawaii"}) \cite{Petroni2019LanguageMA}. 
However, such implicit query-based knowledge falls short of many desirable properties of a full-scale KG such as ConceptNet \cite{alkhamissi2022review}, including easy access, browsing, or even editing \cite{Zhu2020ModifyingMI, decao2021editing}, as well as assurance of knowledge quality thanks to the symbolic nature \cite{anderson2020neurosymbolic}. Symbolic Knowledge Distillation \cite[SKD, ][]{west-etal-2022-symbolic} explicitly extracts a knowledge base from GPT-3. However, the approach exclusively relies on the strong in-context learning capability of GPT-3 and thus is not applicable to other rich LMs such as \textsc{Bert} \cite{Devlin2019BERTPO} and \textsc{RoBERTa} \cite{Liu2019RoBERTaAR}. Moreover, its use of a quality discriminator trained on existing KGs can limit its generalization to new relations not included in the training data.

\footnotetext[1]{"NER", "CR", "RE" refer to "named entity recognition", "coreference resolution", "relation extraction", respectively.}
\footnotetext[2]{SKD has an optional filter that requires existing KG to finetune, which doesn't work for arbitrary relations.}


In this paper, we propose a new approach of harvesting massive KGs of arbitrary new relations from any pretrained LMs. Given minimal user input of a relation definition, including a prompt and a few shot of example entity pairs, our approach automatically searches within the LM to extract an extensive set of high-quality knowledge about the desired relation. To ensure search efficiency in the vast space of entity pairs, we devise an effective search-and-rescore strategy. We also adapt the previous prompt paraphrasing mechanism \cite{Jiang2020HowCW, Newman2021PAdaptersRE} and enhance with our new rescore strategy for prompt weighting, leading to consistent and accurate outcome knowledge. 

We apply our approach on a range of LMs of varying capacities, such as \textsc{RoBERTa}, \textsc{Bert}, and \textsc{DistilBert}. In particular, we harvest knoweldge of over 400 new relations (an order of magnitude more than ConceptNet relations) not available in preexisting KGs and previous extraction methods. Extensive human and automatic evaluations show our approach successfully extracts diverse accurate knowledge, including tuples for complex relations such as \texttt{``A is capable of, but not good at, B''} and 3-ary relations such as \texttt{``A can do B at C''}. Interestingly, the resulting KGs also serve as a symbolic interpretation of the source LMs, revealing new insights into their knowledge capacities in terms of varying factors such as model size, pretraining strategies, and distillation.

\section{Related Work}

\paragraph{Knowledge graph construction} 
\label{sec:relatedworks}
Popular knowledge bases or KGs are usually constructed with heavy human labor. For example, 
WordNet \citep{Fellbaum2000WordNetA} is a lexical database that links words into semantic relations;
ConceptNet \citep{Speer2017ConceptNet5A} is a large commonsense knowledge graph presented as a set of knowledge triples; ATOMIC \citep{Sap2019ATOMICAA} is a crowd-sourced social commonsense KG of if-then statements. Recently, Automatic Knowledge Base Construction (AKBC) as a research focus has led to various approaches (summarized in Table~\ref{tab:summary}). Text mining-based works aim for knowledge extraction from text. A typical information extraction system \citep{Angeli2015LeveragingLS} is composed of several sub-tasks like coreference resolution, named entity recognition, and relationship extraction. Some works on commonsense knowledge extraction include WebChild \citep{tandon2014webchild}, TransOMCS \citep{Zhang2020TransOMCSFL}, DISCOS \citep{fang2021discos}, Quasimodo \citep{romero2019commonsense}, ASCENT \citep{nguyen2021refined}. These extraction pipelines are based on linguistic pattern, and involve complex engineering such as corpus selection, term aggregation, filtering, etc. Recent attempts also utilize LMs for AKBC. \citealt{Wang2021StructureAugmentedTR} finetuned LMs for link prediction. \citealt{Feldman2019CommonsenseKM, Bouraoui2020InducingRK} utilized LMs to score entity pairs collected from the Internet or missing edges in existing KGs. COMET \citep{Bosselut2019COMETCT} is a generative LM trained to predict tail entities given head entities and relations. \citealt{west2021symbolic} distill the knowledge in GPT-3 to a generative LM. By prompting GPT-3 \citep{Brown2020LanguageMA} with examples, they produced \textsc{ATOMIC$_{10x}$} to teach the student model. Yet, this method requires the strong few-shot learning ability of GPT-3 and is not generally applicable to most LMs. To the best of our knowledge, our framework is the first to construct a KG by extracting purely from an LM (with the minimal definition of relations as input). The new paradigm can also be seen as optimizing a symbolic KG with (pretrained) neural models as supervision \cite{Hu2022Toward}, which inverts the conventional problem of using symbolic knowledge to learn neural networks \cite{hu2016harnessing}.


\paragraph{LMs as knowledge bases}
Another line of works attempted to use LMs as knowledge bases (LAMA, \citealt{Petroni2019LanguageMA}). These works are also known as factual probing because they measured how much knowledge is encoded in LMs. This is usually implemented by prompting methods and leveraging the masked LM pretraining task. LPAQA \citep{Jiang2020HowCW} proposes to use text mining and paraphrasing to find and select prompts to optimize the prediction of a single or a few correct tail entities, instead of extensively predicting all the valid entity pairs like in our framework. AutoPrompt \citep{Shin2020ElicitingKF}, \citealp{Qin2021LearningHT} and OPTIPrompt \citep{Zhong2021FactualP} learn discrete or continuous prompts automatically with an additional training set. Though making prompts unreadable, these methods achieve higher accuracy on the knowledge probing tasks. Our framework differs from these works in that we aim to explicitly harvest knowledge graphs instead of measuring the knowledge in a simplified setting.




\paragraph{Consistency of LMs} Consistency is a significant challenge for LMs, which stresses that they should not produce conflicting predictions across inference sessions. For example, models should behave invariantly under inputs with different surface forms but the same meaning. \citealt{elazar2021measuring} analyzed the consistency of pretrained LMs with respect to factual knowledge. \citealt{Jiang2020HowCW} used paraphrasing to improve factual probing. \citealt{Newman2021PAdaptersRE} trains an additional layer on top of word embedding to improve consistency. Recently, consistency is also shown helpful to improve the reasoning ability of large LMs \cite{wang-etal-2022-self, jung2022maieutic, hao2023reasoning}. 
In our framework, the extracted entity pairs for each relation are enforced to consistently satisfy a diverse set of prompts and regularized by several scoring terms.
 

\section{Harvesting KGs from LMs}
\label{sec:methods}

\begin{figure*}
    \centering
    \includegraphics[width=0.95\textwidth]{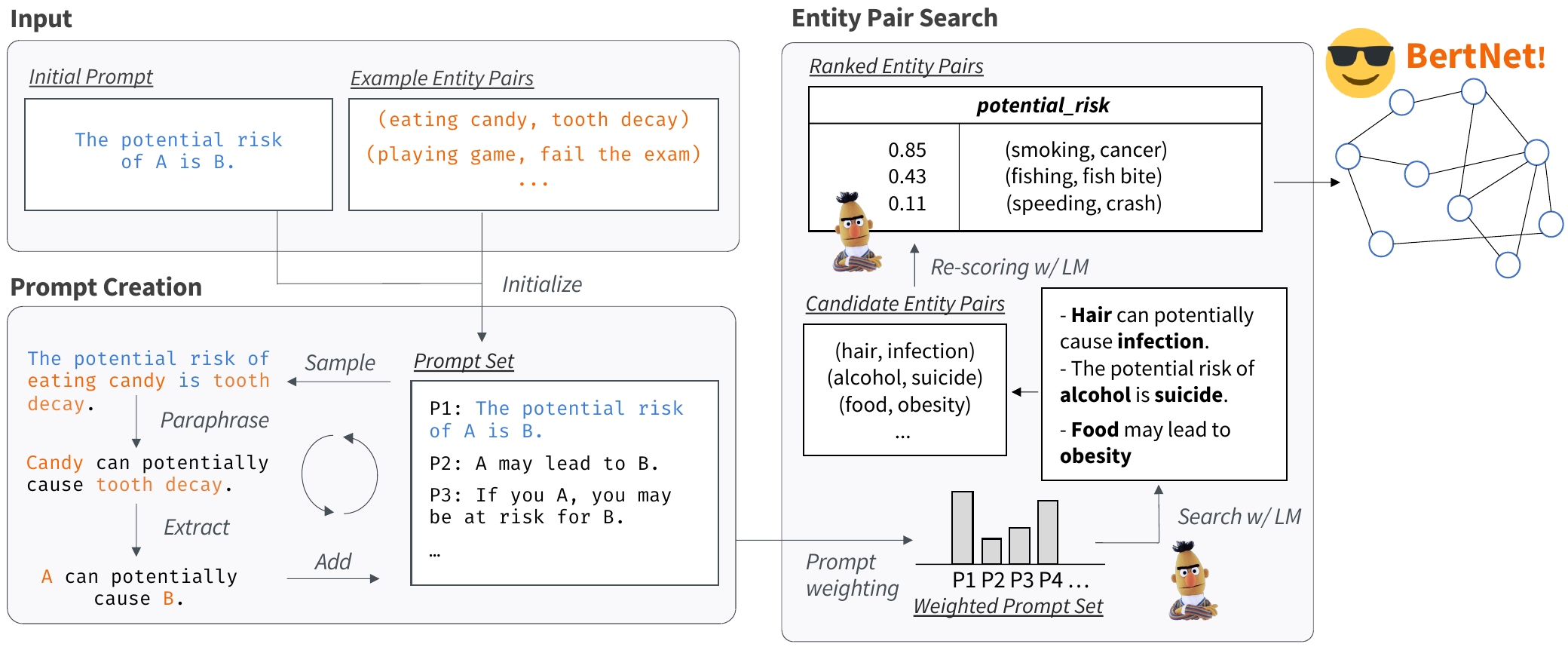}
    \caption{An overview of the knowledge harvesting framework. Given the minimal definition of the relation as input (an initial prompt and a few shot of example entity pairs), the approach first automatically creates a set of prompts expressing the relation in a diverse ways (\S\ref{sec:search_prompts}). The prompts are weighted with confidence scores. We then use the LM to search a large collection of candidate entity pairs, followed by re-scoring/ranking that yields the top entity pairs as the output knowledge (\S\ref{sec:search_ep}). 
    }
    \label{fig:framework}
    \vspace{-5pt}
\end{figure*}

This section presents the proposed framework for extracting a relational KG from a given pretrained LM, where the LM can be arbitrary fill-in-the-blank models such as \textsc{Bert} \citep{Devlin2019BERTPO}, \textsc{RoBertA} \citep{Liu2019RoBERTaAR}, \textsc{BART} \citep{Lewis2020BARTDS}, or \textsc{GPT-3} (with appropriate instructions) \citep{Brown2020LanguageMA}. The KG consists of a set of knowledge tuples in the form \tuple{head entity ($h$), relation ($r$), tail entity ($t$)}.
Our approach utilizes the LM to automatically harvest a large number of appropriate entity pairs $(h_1, t_1), (h_2, t_2), \ldots$, for every given relation $r$. This presents a more challenging problem than traditional LM probing tasks, which typically predict a single tail entity or a small number of valid tail entities given a head entity and relation. 

Our approach for extracting knowledge tuples of a specific relation of interest, such as \texttt{"potential\_risk"} as depicted in Figure~\ref{fig:framework}, only requires minimal input information that defines the relation. This includes an initial prompt, such as \texttt{"The potential risk of A is B"} and a small number of example entity pairs, such as \tuple{eating candy, tooth decay}. The prompt provides the overall semantics of the relation, while the example entity pairs clarify possible ambiguities. For new relations not included in existing KGs, it is impractical to require a large set (e.g., hundreds) of example entity pairs as in previous knowledge probing or prompt optimization methods \cite{Petroni2019LanguageMA, Jiang2020HowCW, Shi2019NeuralLN, Zhong2021FactualP}. In contrast, our approach necessitates only a small number of example entity pairs, for example, as few as 2 in our experiments, which can easily be collected or written by users.

In the following sections, we describe the core components of our approach, namely the automatic creation of diverse prompts with confidence weights (\S\ref{sec:search_prompts}) and the efficient search to discover consistent entity pairs (\S\ref{sec:search_ep}) that compose the desired KGs. Figure~\ref{fig:framework} illustrate the overall framework.

\subsection{Creating Diverse Weighted Prompts}
\label{sec:search_prompts}

Our automated approach utilizes input information, specifically the initial prompt and several example entity pairs, to generate a set of semantically consistent but linguistically diverse prompts for describing the relation of interest. The generated prompts are assigned confidence weights to accurately measure consistency of knowledge in the subsequent step (\S\ref{sec:search_ep}).

To generate diverse prompts for a desired relation, we begin by randomly selecting an entity pair from a example set and inserting it into an initial prompt to form a complete sentence. This sentence is then passed through an off-the-shelf text paraphrase model, which produces multiple paraphrased sentences with the same meaning. By removing the entity names, each paraphrased sentence results in a new prompt that describes the desired relation. To ensure a wide range of expressions of the relation, we retain only those prompts that are distinct from one another in terms of edit distance. This process is repeated by continuously paraphrasing the newly created prompts until a minimum of 10 prompts for the relation have been collected.

The automatic generation of prompts can be imprecise, resulting in prompts that do not accurately convey the intended relation. To mitigate this, we propose a reweighting method that utilizes compatibility scores to calibrate the impact of each prompt in the subsequent knowledge search step. Specifically, we evaluate the compatibility of new prompts with example entity pairs by measuring the likelihood of the prompts under a LM, considering both the individual entities and the entity pair as a whole. This allows us to determine the appropriate weights for each prompt and improve the precision of the knowledge search process. Formally, the compatibility score between an entity pair $(h, t)$ and a prompt $p$ can be written as:
\begin{equation}
\small
\label{eqn:pem}
\begin{aligned}
    f_{\text{LM}}(\langle & h, t \rangle, p) = \alpha \log P_{\text{LM}}( h, t \mid p ) \\
    & + (1 - \alpha) \min \left\{ \log P_{\text {LM}}(h \mid p), \log P_{\text {LM}}(t \mid p, h) \right\}
\end{aligned}
\end{equation}
where the first term is the joint log-likelihood under the LM distribution $P_{LM}$, the second term is the minimum individual log-likelihood given the prompt (and the other entity), and $\alpha$ is a balancing factor ($\alpha=2/3$ in our experiments). We compute the average compatibility score of each created prompt over all example entity pairs, and the weight of the prompt is then defined as the softmax-normalized score across all prompts.

\begin{figure*}[h]
    \centering
    \includegraphics[width=0.95\textwidth]{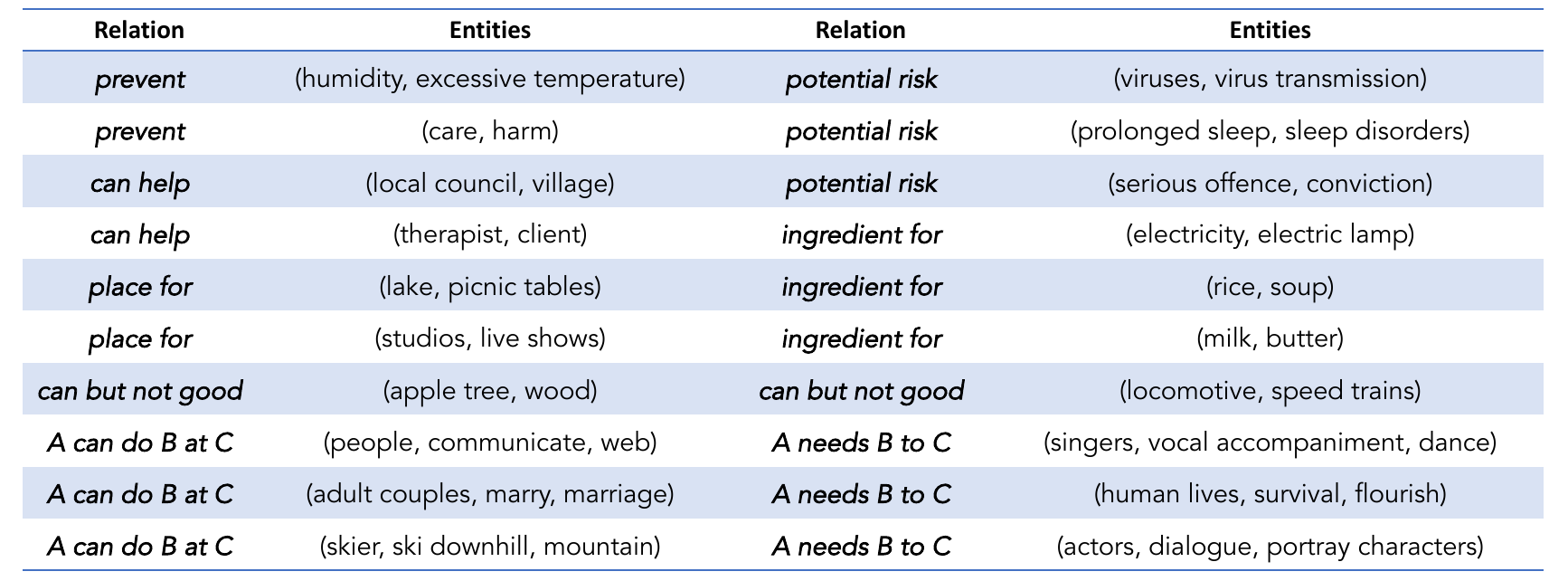}
    \caption{Examples of knowledge tuples harvested from \textsc{DistillBert} (randomly sampled). The first 7 rows shows relations with two entities (head and tail), and last 3 rows shows more complex relations with 3 entities.}
    \label{fig:examples}
\end{figure*}

\subsection{Efficient Search for Consistent Knowledge}
\label{sec:search_ep}

With the set of prompts and corresponding confidence weights obtained in the steps described in Section \ref{sec:search_prompts}, we proceed to search entity pairs that consistently align with all prompts. To guide the searching process and evaluate the compatibility of searched-out entity pairs $( h^{new}, t^{new})$, we reuse the previously defined prompt/entity-pair compatibility function (Eq.\ref{eqn:pem}), and intuitively define consistency as the weighted average of its compatibility with the various prompts, i.e.,
\begin{equation}
\small
    \text{consistency}(( h^{\text{new}}, t^{\text{new}})) = \sum\nolimits_{p} w_{p} \cdot f_{\text{LM}}(( h^{\text{new}}, t^{\text{new}}), p)
\label{eqn:consistent}
\end{equation}
where $w_p$ is the prompt weight and the sum is over all automatically created prompts as above, so that entity pairs compatible with all prompts are considered to be consistent.

Based on the consistency criterion, we develop an efficient search strategy to search for consistent entity pairs. A straightforward approach involves enumerating all possible pairs of entities, calculating their respective consistency scores, and selecting the top-K entity pairs with the highest scores as the resulting knowledge. However, this approach can be computationally expensive due to the large vocabulary size $V$ (e.g., $V=50,265$ for \textsc{RoBerta}) and the high time complexity of the enumeration process (i.e., $O(V^2)$ even when each entity consists of only one token). To overcome this limitation, we have proposed an appropriate approximation that leads to a more efficient \textit{search and re-scoring} method. Specifically, we first use the minimum individual log-likelihoods (i.e., the second term in the compatibility score Eq.\ref{eqn:pem}) weighted averaged across different prompts (similar as in Eq.\ref{eqn:consistent}), to propose a large set of candidate entity pairs. The use of the minimum individual log-likelihoods allows us to apply pruning strategies, such as maintaining a heap and eliminating entities ranked outside top-K in every single searching step. Once we have collected a large number of proposals, we re-rank them using the full consistency score in Eq.\ref{eqn:consistent} and select the top-K instances as the output knowledge. We describe more nuanced handling in the search procedure (e.g., the processing of multi-token entities, detailed pruning strategies) in the appendix. 

\paragraph{Generalization to complex relations}
\label{sec:complex} 
Most existing KGs or knowledge bases include relations that are predicates connecting two entities, e.g., \texttt{"A is capable of B"}. However, many real-life relations are more complex. Our approach is flexible and easily extensible to extract knowledge about these complex relations. We demonstrate this in our experiments by exploring two cases: (1) \textbf{\textit{highly customized relations}} that have specific and sophisticated meanings, such as \texttt{"A is capable of, but not good at, B"}. This type of sophisticated knowledge is often difficult for humans to write down on a large scale. Our automatic approach naturally supports harvesting this kind of knowledge given only an initial prompt and a few example entities that can be collected easily, e.g., \tuple{dog, swim}, \tuple{chicken, fly}, etc.; (2) \textbf{\textit{N-ary relations}} involving more than two entities, such as \texttt{"A can do B at C"}. Our approach can straightforwardly be extended to handle $n$-ary relations by generalizing the compatibility score and search strategy accordingly to accommodate more than two entities.

\begin{table*}[]
\small
    \centering
\begin{tabular}{@{}crcccc@{}}
\toprule
Paradigm & Method (Size) & Relation Set & \#Relations & Accuracy (\%) & Novelty (\%) \\

\midrule
\vspace{0.1em}
\multirow{10}{*}{Ours} & RobertaNet\textsuperscript{} (122.2k) & \cellcolor{yellow!25}{Auto} & \cellcolor{yellow!25}{487} & 65.3& -  \\ \cmidrule{2-6}
&RobertaNet\textsuperscript{} (2.2K) & Human & 12 & 81.8 & - \\
&RobertaNet\textsuperscript{} (7.3K) & Human & 12 & 68.6  & - \\
&RobertaNet\textsuperscript{} (23.6k) & Human & 12 & 58.6 & - \\ \cmidrule{2-6}
&RobertaNet\textsuperscript{} (6.7K) & ConceptNet & 20 & 88.0& 64.4 \\
&RobertaNet\textsuperscript{} (24.3K) & ConceptNet & 20 & 81.6 & 68.8 \\
&RobertaNet\textsuperscript{} (230K) & ConceptNet & 20 & 55.0  & 87.0 \\ 
\midrule\midrule
\multirow{2}{*}{KG Completion} & COMET\textsuperscript{} (6.7K) & ConceptNet & 20 & 92.0 & 35.5 \\
& COMET\textsuperscript{} (230K) & ConceptNet & 20 & 66.6 & 72.4 \\
\midrule
\multirow{3}{*}{Text Mining} & WebChild (4.6M) & - & 20 & 82.0*&- \\
&ASCENT (8.6M) & - & - & 79.2*&- \\
&TransOMCS (18.4M) & ConceptNet & 20 & 56.0* &98.3 \\
\bottomrule
\vspace{-5pt}
\end{tabular}
\vspace{-10pt}
    \caption{Statistics of KGs constructed with different methods. Different paradigms of works \textbf{can not be directly compared} due to their different settings discussed in Table~\ref{tab:summary}. We put the results together for reference purpose.
    \textit{Novelty} refers to the proportion of entities that do not appear in ConceptNet, so only the methods with ConceptNet relations set have  \textit{Novelty} numbers. The accuracy with $^*$ are from the original papers and subject to different evaluation protocol.
    As a finetuned knowledge base completion model, COMET\cite{Bosselut2019COMETCT} can only predict the tail entity given a source entity and a relation, we generate KGs with COMET by feeding it the head entity produced by our \textsc{RobertaNet}. The bottom block of the table summarizes the results from some major text mining methods described in Table \ref{tab:summary}, including WebChild \cite{tandon2014webchild}, ASCENT \cite{nguyen2021refined} and TransOMCS \cite{Zhang2020TransOMCSFL}.
    }
    \label{tab:stats}
    \vspace{-5pt}
\end{table*}

\paragraph{Symbolic interpretation of neural LMs}

The harvested knowledge tuples, as consistently recognized across varying prompts by the LM, can be considered as the underlying "beliefs" of the LM about the world \cite{stich1979animals,Hase2021DoLM}. These fully symbolic and interpretable tuples provide a means for easily browsing and analyzing the knowledge capabilities of the black-box LM. For example, via these outcome KGs, one can compare different LMs to understand the performance impact of diverse configurations, such as model sizes and pretraining strategies, as demonstrated in our experiments.

\section{Experiments}

To evaluate our framework, we extract knowledge of diverse new relations from various language models, and conduct human evaluation. We then make deeper analysis of prompt creation and scoring function in our framework. Finally, by utilizing our framework as a tool to interpret the knowledge stored in language models, we have made noteworthy observations regarding the knowledge capacity of black-box models. 

\vspace{-5pt}


\subsection{Setup}
\label{sec:setup}
\paragraph{Relations} 
We evaluate our framework with several relation sets: (1) \textbf{ConceptNet} \cite{Speer2017ConceptNet5A}: Following \citealt{li-etal-2016-commonsense}, we filter the KG and use a set of 20 common relations (e.g. \textsc{has\_subevent}, \textsc{motivated\_by\_goal}). The initial prompts for these relations are from the ConceptNet repository, and we randomly sample 5 example entity
pairs from the ConceptNet KG for each relation. (2) \textbf{LAMA} \cite{Petroni2019LanguageMA}: Following previous works,
we use the T-REx split (41 relations from WikiPedia, such as capital\_of, member\_of). For each relation,
the human-written prompt provided in \citealt{Petroni2019LanguageMA} is used as the initial prompt and we randomly sample 5 example entity pairs for each relation. (3) \textbf{Human}: We write 12 new relations of interests that can hardly be found in any existing KGs, and manually write an initial prompt and 5 example entity pairs for them. The resulting relations include complex relations as described in Section~\ref{sec:complex}. (4) \textbf{Auto}: Besides relations from existing KGs and human-written ones, we automatically derive a large set of relations from E-KAR \cite{chen2022kar}, a dataset for analogical reasoning. In the original dataset, given an entity pair, e.g. \tuple{id\_card, identity}, the task is to select an analogous tuple from multiple choices, e.g. \tuple{practice license, qualification}. To turn a sample in E-KAR into a relation, we use the tuple in the question and the correct choices as 2 example entity pairs, and extract the initial prompt from the explanation provided in E-KAR (e.g. \textit{Proof of A requires B.}), resulting in 487 relations. Some of the relations are not straightforward, making this relation set more difficult than other ones. \footnote{For reference, finetuned \textsc{RoBERTa-large} achieves about 50\% accuracy on the original dataset.}

\label{para:hparam}
\subsection{Extracting Knowledge of Diverse New Relations}

Our framework is applied to extract knowledge graphs from LMs with relations of ConceptNet, Auto, and Human. The accuracy of the extracted knowledge is then evaluated with human annotation using Amazon Mechanical Turk (MTurk). Each extracted knowledge tuple is labeled for correctness by three annotators using a True/False/Unjudgeable judge. A tuple is considered "accepted" if at least two annotators deem it to be true knowledge, and "rejected" if at least two annotators rate it as false. Here we refers portion of accepted tuples as accuracy. 

The statistics of our resulting KGs are listed in Table \ref{tab:stats}. Besides, we also put the results of other paradigms of methods, including COMET for KG completion and text-mining based methods (Figure~\ref{fig:paradigms}). Note that the results across different paradigms are generally not directly comparable due to vastly different settings. Yet we still collect the results together for reference purpose.
From our RebertaNet with relation set "Auto", we are able to extract a reasonably large sets of knowledge (122K), by extracting knowledge with 487 easy-to-collect "Auto" relations. The set of relation is an order of magnitude larger than the predefined set of relations in both KG completion and text mining based on ConceptNet as shown in the table.
The accuracy of 65\% is at a comparable level with that of COMET (230K) and TransOMCS (18.4M), which is reasonable especially considering our method solely uses an LM as the source of knowledge without any external training data, bringing flexibility to dynamically incorporate new relations. 
Besides, for our RobertaNet on ConceptNet relations, although the numbers listed in the table are not simply comparable, we can still find that RobertaNet achieves similar accuracy and absolutely higher novelty comparing with the knowledge from COMET, which is already finetuned using large number of knowledge terms under the same set of ConceptNet relations.
Further, our results on the "human" relation set demonstrate that our RobertaNet keeps working comfortably on our highly realistic relations of user interests, including the complex ones as described in section \S\ref{sec:complex}. 
\begin{figure}[t]
\small
\centering
\begin{tabular}{@{}r c c}
\toprule
Methods & Acc  & Rej \\ \midrule
\textsc{Autoprompt}           & 0.33     &  0.47  \\
\textsc{Human Prompt} & 0.60        &0.27   \\
\textsc{Top-1 Prompt} (Ours)    & 0.69   &  0.23 \\
\textsc{Multi Prompts} (Ours) &\textbf{0.73} & \textbf{0.20}  
\\ \bottomrule
\end{tabular}
\captionof{table}{The portions of accepted and rejected tuples in human evaluation across settings, with the \textsc{RoBERTa-large} as the LM.}
\label{tab:human_setting}
\vspace{-10pt}
\end{figure}
We showcase knowledge samples harvested from \textsc{DistillBERT} in Figure \ref{fig:examples}.

\subsection{Analyzing Automatic Prompt Creation}
\label{sec:eval_prompts}

To evaluate the effect of our automatic creation of prompts, we compare the generated KGs under several settings on the Human relations: (1) \textbf{Multi-Prompts} refers to the the full framework described in \S\ref{sec:methods} which use the automatically created diverse prompts in knowledge search. (2) \textbf{Top-1 Prompt}: To ablate the effect of ensembling multiple prompts, we evaluate the variant that uses only the prompt with largest weight (\S\ref{sec:search_prompts}) for knowledge extraction.
(3) \textbf{Human Prompt}: To further understand the effectiveness of the automatically created prompts, we assess the variant that uses the initial prompt of each relation. (4) \textsc{\bf AutoPrompt} \cite{Shin2020ElicitingKF}, which was proposed to learn prompts by optimizing the likelihood of tail entity prediction on the training set. To fit in our setting, we adapt it to optimize the compatibility score (Eq.\ref{eqn:pem}) on the example entity pairs. 
We omit other prompt tuning work  \cite[e.g.,][]{Zhong2021FactualP, Qin2021LearningHT} because they either are difficult to fit in our problem or require more training data and fail with only the few shot of example entity pairs in our setting.

\begin{table}[t]
\small
\centering
\begin{tabular}{@{}r c c}
\toprule
Source LMs         &Acc & Rej \\ \midrule
\textsc{DistilBert}    & 0.67    & 0.24     \\
\textsc{Bert-base}    & 0.63    & 0.26     \\
\textsc{Bert-large}  & 0.70    & 0.22     \\
\textsc{RoBERTa-base} & 0.70    & 0.22     \\
\textsc{RoBERTa-large}& 0.73    & 0.20     \\ \bottomrule
\end{tabular}
\captionof{table}{The portions of accepted and rejected tuples in human evaluation across different LMs, using the \textsc{Multi-Prompts} approach.}
\label{tab:human_model}
\vspace{-5pt}
\end{table}

We harvest 1000 tuples for each Human relation, and evaluate them with human annotation. The annotation results are presented in Table \ref{tab:human_setting} (We also list the detailed results per relation in Table~\ref{tab:results} for reference)
Our \textsc{Top-1 Prompt} significantly improves the accuracy up to 9\% over the \textsc{Human Prompt}, demonstrating the effectiveness of our prompt searching algorithm in generating high-quality prompts. \textsc{Multi-Prompts} further improves the accuracy by an additional 4\%, indicating that the combination of diverse prompts better captures the semantics of a relation. However, the method utilizing the optimized prompt by \textsc{AutoPrompt} results in lower accuracy than the use of human or searched prompts. This can be attributed to the insufficient number of example knowledge tuples used to learn effective prompts for the desired relations.

Based on the results above, we move a step forward to see how the created prompts influence the subsequent scoring module in the framework. Specifically, we study both the precision and recall of our scoring function parameterized by the prompts, to see if the automatically created prompts (\S\ref{sec:search_prompts}) bring the consistency scoring (\S\ref{sec:search_ep}) better balance of knowledge accuracy (precision) and coverage (recall). To compare with other scoring methods that are restricted to specific sets of relations, this experiment was conducted using existing terms from both the ConceptNet and LAMA datasets.

Specifically, we use the knowledge tuples from ConceptNet and LAMA as positive samples (\S\ref{sec:setup}), and synthesize the same amount of negative samples with the same strategy in \citet{li-etal-2016-commonsense} by random replacing entities or relations in a true knowledge tuple. Each scoring function ranks the samples based on the scores from high to low. We can then compute both the \emph{precision} and \emph{recall} of positive samples at different cut-off points along the ranking, and plot the precision-recall curves for each method.

The automatic evaluation setting on given knowledge terms enables us to adapt existing prevalent works, e.g., KG completion and factual probing (Table~\ref{tab:summary}), for comparison with our approach:
{\bf (1)} \textsc{\bf COMET} \cite{Bosselut2019COMETCT} is a transformer-based KG completion model trained to predict the tail entity $t$ conditioning on the head entity and relation $(h, r)$ on ConceptNet. We use its log-likelihood $\log P(t|h, r)$ as the score for each given knowledge tuple.
{\bf (2)} \textsc{\bf LPAQA} \cite{Jiang2020HowCW} collects a set of prompts on LAMA with text mining and paraphrasing, and optimize their weights towards the objective of $\log P(t|h, r)$ on training samples.

\begin{figure}[t]
\centering
\includegraphics[width=0.45\textwidth]{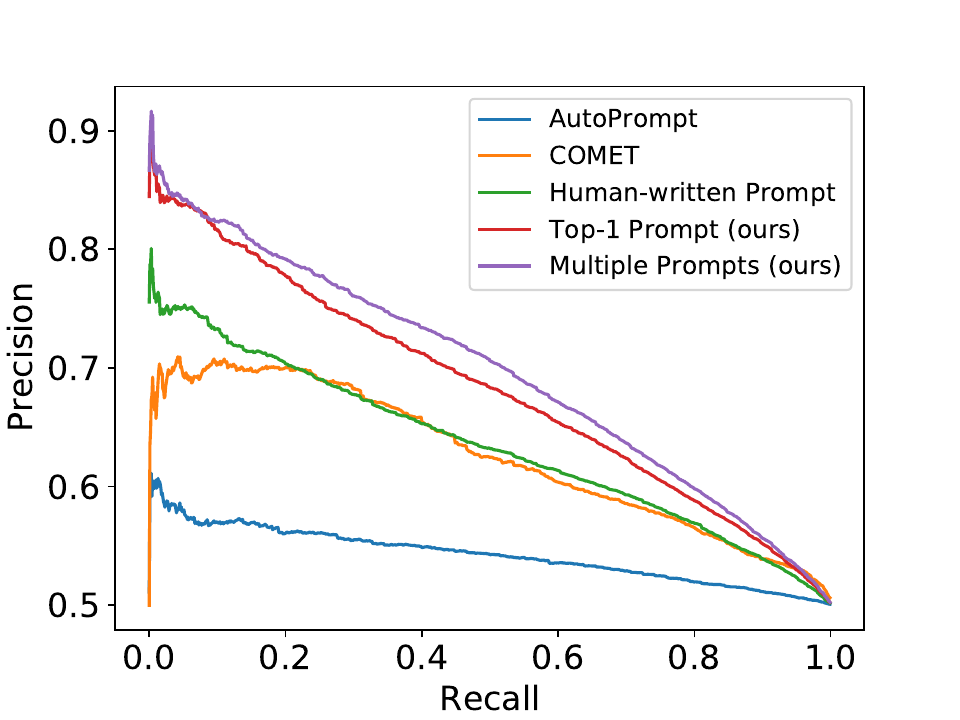}
\captionof{figure}{Precision-recall on ConceptNet relations.}
\label{fig:curve_cn}
\vspace{-10pt}
\end{figure}

\begin{figure}[t]
\centering
\includegraphics[width=0.45\textwidth]{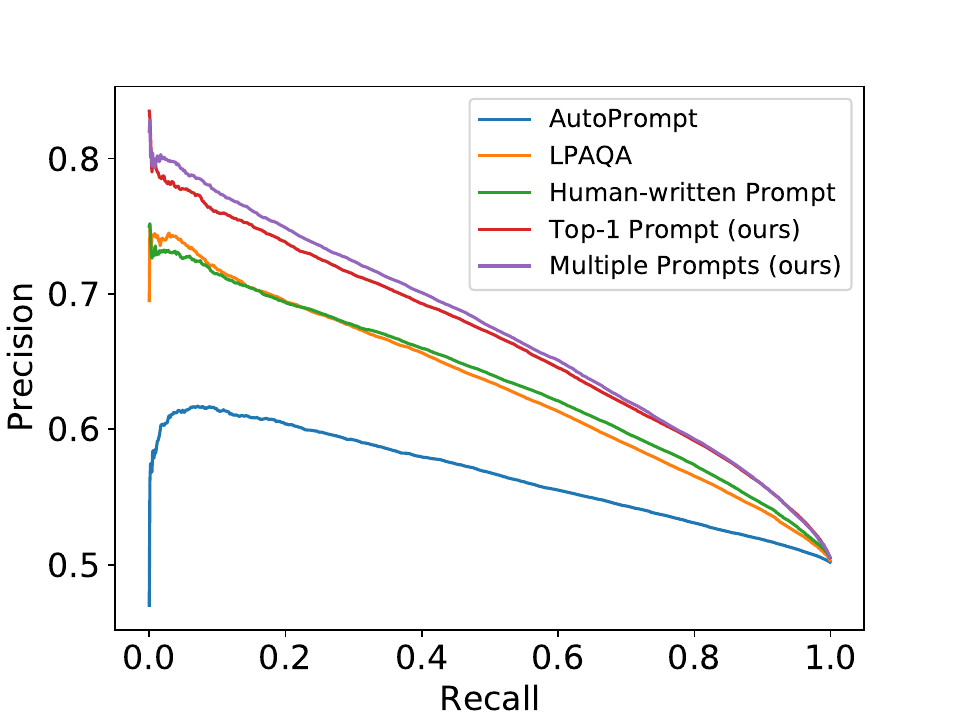}
\captionof{figure}{Precision-recall curve on LAMA relations. 
}
\label{fig:curve_lama}
\vspace{-10pt}
\end{figure}

The resulting precision-recall curves on ConceptNet and LAMA knowledge are shown in Figure~\ref{fig:curve_cn} and Figure~\ref{fig:curve_lama}, respectively. Scoring with multiple prompts always achieves best performance, followed by Top-1 prompts and then Human-written prompts. The finding is consistent with previous experiments, which verified the effectiveness of our scoring function design. Our framework also outperforms other baselines, such as \textsc{Comet} on ConceptNet and \textsc{LPAQA} on LAMA. Though trained with labeled data, these methods are only optimized to completing a tail entity given a query, in stead of scoring an entity pair, which is essential to extract KGs from LMs. 

\subsection{Analysis of Knowledge in Different LMs} 
\label{sec:x-net}
As previously mentioned in Section \S\ref{sec:methods}, the resulting knowledge graphs can be viewed as a symbolic interpretation of LMs. We extract knowledge graphs from 5 distinct language models and submit them to human annotation evaluation. The findings are presented in Table \ref{tab:human_model} (The detailed results per relation is listed in Table~\ref{tab:results}), which sheds some new light on several knowledge-related questions regarding the LMs' knowledge capacity.

{\bf Does a larger LM encode better knowledge?} The large version of BERT and RoBERTa have the same pretraining corpus and tasks as their base versions, but have larger model architecture in terms of layers (24 v.s. 12), attention heads (16 v.s. 12), and the number of parameters (340M v.s. 110M). We can see that the accuracies of BertNet-large and RoBERTaNet-large are around 7\% and 3\% higher than their base version, separately, indicating the larger models indeed encoded better knowledge than the base models.

{\bf Does better pretraining bring better knowledge?} RoBERTa uses the same architecture as BERT but with better pretraining strategies, like dynamic masking, larger batch size, etc. In their corresponding KGs from our framework, RoBERTaNet-large performs better than BertNet-large (0.73 v.s. 0.70), and RoBERTaNet-base is also better than BertNet-base (0.70 v.s. 0.63), showing that the better pretraining in RoBERTa leads to better knowledge learning and storage.

{\bf Is knowledge really kept in the knowledge distillation process?} DistilBERT is trained by distilling BERT-base, and it reduces 40\% parameters from the latter. Interestingly, the knowledge distillation process instead improves around 4\% of accuracy in the result knowledge graph. This should be attributed to the knowledge distillation process which might eliminate some noisy information from the teacher model.

\section{Conclusion}
We have developed an automatic framework that extracts a KG from a pretrained LM (e.g, \textsc{Bert}, \textsc{RoBerta}), in an efficient and scalable way, resulting in a family of new KGs, which we refer to as \textsc{BertNet}, \textsc{RoBertaNet}, etc.
Our framework is capable of extracting knowledge of arbitrary new relation types and entities, without being restricted by pre-existing knowledge or corpora. The resulting KGs also serve as interpretation of source LMs.

\paragraph{Limitations} Our current design and experimental studies are limited on LMs in the generic domain, and are not yet been studied in specific domains such as extracting healthcare knowledge from relevant neural models. We leave the exciting work of harvesting knowledge from various kinds of neural networks across applications and domains in the future work.

\paragraph{Ethical considerations} In this work, the harvested knowledge is automatically generated by LMs. We would like to note that the language models could possibly generate unethical knowledge tuples, same with the risks of other applications using language models for generation. 
We hope that the knowledge extraction study could offer techniques to better interpret and understand the language models, and in turn foster the future research of language model ethics.
Since the knowledge graph only consists simple phrases, we think filtering sensitive words would be effective.
No foreseeable negative societal impacts are caused by the method itself.

\bibliography{anthology,custom}

\begin{thebibliography}{53}
\expandafter\ifx\csname natexlab\endcsname\relax\def\natexlab#1{#1}\fi

\bibitem[{AlKhamissi et~al.(2022)AlKhamissi, Li, Celikyilmaz, Diab, and
  Ghazvininejad}]{alkhamissi2022review}
Badr AlKhamissi, Millicent Li, Asli Celikyilmaz, Mona Diab, and Marjan
  Ghazvininejad. 2022.
\newblock A review on language models as knowledge bases.
\newblock \emph{arXiv preprint arXiv:2204.06031}.

\bibitem[{Anderson et~al.(2020)Anderson, Verma, Dillig, and
  Chaudhuri}]{anderson2020neurosymbolic}
Greg Anderson, Abhinav Verma, Isil Dillig, and Swarat Chaudhuri. 2020.
\newblock Neurosymbolic reinforcement learning with formally verified
  exploration.
\newblock \emph{Advances in neural information processing systems},
  33:6172--6183.

\bibitem[{Angeli et~al.(2015)Angeli, Johnson, and
  Manning}]{Angeli2015LeveragingLS}
Gabor Angeli, Melvin Johnson, and Christopher~D. Manning. 2015.
\newblock Leveraging linguistic structure for open domain information
  extraction.
\newblock In \emph{ACL}.

\bibitem[{Bordes et~al.(2013)Bordes, Usunier, Garcia-Duran, Weston, and
  Yakhnenko}]{bordes2013translating}
Antoine Bordes, Nicolas Usunier, Alberto Garcia-Duran, Jason Weston, and Oksana
  Yakhnenko. 2013.
\newblock Translating embeddings for modeling multi-relational data.
\newblock \emph{Advances in neural information processing systems}, 26.

\bibitem[{Bosselut et~al.(2019)Bosselut, Rashkin, Sap, Malaviya, Çelikyilmaz,
  and Choi}]{Bosselut2019COMETCT}
Antoine Bosselut, Hannah Rashkin, Maarten Sap, Chaitanya Malaviya, Asli
  Çelikyilmaz, and Yejin Choi. 2019.
\newblock Comet: Commonsense transformers for knowledge graph construction.
\newblock \emph{The Association for Computational Linguistics}.

\bibitem[{Bouraoui et~al.(2020)Bouraoui, Camacho-Collados, and
  Schockaert}]{Bouraoui2020InducingRK}
Zied Bouraoui, Jos{\'e} Camacho-Collados, and Steven Schockaert. 2020.
\newblock Inducing relational knowledge from bert.
\newblock In \emph{AAAI}.

\bibitem[{Brown et~al.(2020)Brown, Mann, Ryder, Subbiah, Kaplan, Dhariwal,
  Neelakantan, Shyam, Sastry, Askell, Agarwal, Herbert-Voss, Krueger, Henighan,
  Child, Ramesh, Ziegler, Wu, Winter, Hesse, Chen, Sigler, Litwin, Gray, Chess,
  Clark, Berner, McCandlish, Radford, Sutskever, and
  Amodei}]{Brown2020LanguageMA}
Tom~B. Brown, Benjamin Mann, Nick Ryder, Melanie Subbiah, Jared Kaplan,
  Prafulla Dhariwal, Arvind Neelakantan, Pranav Shyam, Girish Sastry, Amanda
  Askell, Sandhini Agarwal, Ariel Herbert-Voss, Gretchen Krueger, T.~J.
  Henighan, Rewon Child, Aditya Ramesh, Daniel~M. Ziegler, Jeff Wu, Clemens
  Winter, Christopher Hesse, Mark Chen, Eric Sigler, Mateusz Litwin, Scott
  Gray, Benjamin Chess, Jack Clark, Christopher Berner, Sam McCandlish, Alec
  Radford, Ilya Sutskever, and Dario Amodei. 2020.
\newblock Language models are few-shot learners.
\newblock \emph{ArXiv}, abs/2005.14165.

\bibitem[{Cao et~al.(2021)Cao, Aziz, and Titov}]{decao2021editing}
Nicola~De Cao, Wilker Aziz, and Ivan Titov. 2021.
\newblock \href {https://arxiv.org/abs/2104.08164} {Editing factual knowledge
  in language models}.

\bibitem[{Chen et~al.(2022)Chen, Xu, Fu, Shi, Li, Zhang, Sun, Li, Xiao, and
  Zhou}]{chen2022kar}
Jiangjie Chen, Rui Xu, Ziquan Fu, Wei Shi, Zhongqiao Li, Xinbo Zhang, Changzhi
  Sun, Lei Li, Yanghua Xiao, and Hao Zhou. 2022.
\newblock E-kar: A benchmark for rationalizing natural language analogical
  reasoning.
\newblock \emph{arXiv preprint arXiv:2203.08480}.

\bibitem[{Devlin et~al.(2019)Devlin, Chang, Lee, and
  Toutanova}]{Devlin2019BERTPO}
Jacob Devlin, Ming-Wei Chang, Kenton Lee, and Kristina Toutanova. 2019.
\newblock Bert: Pre-training of deep bidirectional transformers for language
  understanding.
\newblock In \emph{NAACL-HLT (1)}.

\bibitem[{Elazar et~al.(2021)Elazar, Kassner, Ravfogel, Ravichander, Hovy,
  Sch{\"u}tze, and Goldberg}]{elazar2021measuring}
Yanai Elazar, Nora Kassner, Shauli Ravfogel, Abhilasha Ravichander, Eduard
  Hovy, Hinrich Sch{\"u}tze, and Yoav Goldberg. 2021.
\newblock Measuring and improving consistency in pretrained language models.
\newblock \emph{Transactions of the Association for Computational Linguistics},
  9:1012--1031.

\bibitem[{Fang et~al.(2021)Fang, Zhang, Wang, Song, and He}]{fang2021discos}
Tianqing Fang, Hongming Zhang, Weiqi Wang, Yangqiu Song, and Bin He. 2021.
\newblock Discos: Bridging the gap between discourse knowledge and commonsense
  knowledge.
\newblock In \emph{Proceedings of the Web Conference 2021}, pages 2648--2659.

\bibitem[{Feldman et~al.(2019)Feldman, Davison, and
  Rush}]{Feldman2019CommonsenseKM}
Joshua Feldman, Joe Davison, and Alexander~M. Rush. 2019.
\newblock Commonsense knowledge mining from pretrained models.
\newblock In \emph{EMNLP}.

\bibitem[{Fellbaum(2000)}]{Fellbaum2000WordNetA}
Christiane~D. Fellbaum. 2000.
\newblock Wordnet : an electronic lexical database.
\newblock \emph{Language}, 76:706.

\bibitem[{Google(2012)}]{googleKG}
Google. 2012.
\newblock \href
  {https://blog.google/products/search/introducing-knowledge-graph-things-not/}
  {Introducing the knowledge graph: things, not strings}.

\bibitem[{Hao et~al.(2023)Hao, Gu, Ma, Hong, Wang, Wang, and
  Hu}]{hao2023reasoning}
Shibo Hao, Yi~Gu, Haodi Ma, Joshua~Jiahua Hong, Zhen Wang, Daisy~Zhe Wang, and
  Zhiting Hu. 2023.
\newblock \href {http://arxiv.org/abs/2305.14992} {Reasoning with language
  model is planning with world model}.

\bibitem[{Hase et~al.(2021)Hase, Diab, Çelikyilmaz, Li, Kozareva, Stoyanov,
  Bansal, and Iyer}]{Hase2021DoLM}
Peter Hase, Mona~T. Diab, Asli Çelikyilmaz, Xian Li, Zornitsa Kozareva,
  Veselin Stoyanov, Mohit Bansal, and Srini Iyer. 2021.
\newblock Do language models have beliefs? methods for detecting, updating, and
  visualizing model beliefs.
\newblock \emph{ArXiv}, abs/2111.13654.

\bibitem[{Hu et~al.(2016)Hu, Ma, Liu, Hovy, and Xing}]{hu2016harnessing}
Zhiting Hu, Xuezhe Ma, Zhengzhong Liu, Eduard~H Hovy, and Eric~P Xing. 2016.
\newblock Harnessing deep neural networks with logic rules.
\newblock In \emph{ACL (1)}.

\bibitem[{Hu and Xing(2022)}]{Hu2022Toward}
Zhiting Hu and Eric~P. Xing. 2022.
\newblock {Toward} a '{Standard} {Model}' of {Machine} {Learning}.
\newblock \emph{Harvard Data Science Review}, 4(4).
\newblock Https://hdsr.mitpress.mit.edu/pub/zkib7xth.

\bibitem[{Jiang et~al.(2020)Jiang, Xu, Araki, and Neubig}]{Jiang2020HowCW}
Zhengbao Jiang, Frank~F. Xu, J.~Araki, and Graham Neubig. 2020.
\newblock How can we know what language models know?
\newblock \emph{TACL}.

\bibitem[{Jung et~al.(2022)Jung, Qin, Welleck, Brahman, Bhagavatula, Bras, and
  Choi}]{jung2022maieutic}
Jaehun Jung, Lianhui Qin, Sean Welleck, Faeze Brahman, Chandra Bhagavatula,
  Ronan~Le Bras, and Yejin Choi. 2022.
\newblock Maieutic prompting: Logically consistent reasoning with recursive
  explanations.
\newblock \emph{arXiv preprint arXiv:2205.11822}.

\bibitem[{Lewis et~al.(2020)Lewis, Liu, Goyal, Ghazvininejad, Mohamed, Levy,
  Stoyanov, and Zettlemoyer}]{Lewis2020BARTDS}
Mike Lewis, Yinhan Liu, Naman Goyal, Marjan Ghazvininejad, Abdelrahman Mohamed,
  Omer Levy, Veselin Stoyanov, and Luke Zettlemoyer. 2020.
\newblock Bart: Denoising sequence-to-sequence pre-training for natural
  language generation, translation, and comprehension.
\newblock In \emph{ACL}.

\bibitem[{Li et~al.(2019)Li, Liang, Hu, and Xing}]{Li2019KnowledgedrivenER}
Christy~Y. Li, Xiaodan Liang, Zhiting Hu, and Eric~P. Xing. 2019.
\newblock Knowledge-driven encode, retrieve, paraphrase for medical image
  report generation.
\newblock In \emph{AAAI}.

\bibitem[{Li et~al.(2016)Li, Taheri, Tu, and Gimpel}]{li-etal-2016-commonsense}
Xiang Li, Aynaz Taheri, Lifu Tu, and Kevin Gimpel. 2016.
\newblock \href {https://doi.org/10.18653/v1/P16-1137} {Commonsense knowledge
  base completion}.
\newblock In \emph{Proceedings of the 54th Annual Meeting of the Association
  for Computational Linguistics (Volume 1: Long Papers)}, pages 1445--1455,
  Berlin, Germany. Association for Computational Linguistics.

\bibitem[{Liu et~al.(2019)Liu, Ott, Goyal, Du, Joshi, Chen, Levy, Lewis,
  Zettlemoyer, and Stoyanov}]{Liu2019RoBERTaAR}
Yinhan Liu, Myle Ott, Naman Goyal, Jingfei Du, Mandar Joshi, Danqi Chen, Omer
  Levy, Mike Lewis, Luke Zettlemoyer, and Veselin Stoyanov. 2019.
\newblock Roberta: A robustly optimized bert pretraining approach.
\newblock \emph{ArXiv}, abs/1907.11692.

\bibitem[{Luan et~al.(2019)Luan, Wadden, He, Shah, Ostendorf, and
  Hajishirzi}]{luan2019general}
Yi~Luan, Dave Wadden, Luheng He, Amy Shah, Mari Ostendorf, and Hannaneh
  Hajishirzi. 2019.
\newblock A general framework for information extraction using dynamic span
  graphs.
\newblock \emph{arXiv preprint arXiv:1904.03296}.

\bibitem[{Newman et~al.(2021)Newman, Choubey, and
  Rajani}]{Newman2021PAdaptersRE}
Benjamin Newman, Prafulla~Kumar Choubey, and Nazneen Rajani. 2021.
\newblock P-adapters: Robustly extracting factual information from language
  models with diverse prompts.
\newblock \emph{ArXiv}, abs/2110.07280.

\bibitem[{Nguyen et~al.(2021)Nguyen, Razniewski, Romero, and
  Weikum}]{nguyen2021refined}
Tuan-Phong Nguyen, Simon Razniewski, Julien Romero, and Gerhard Weikum. 2021.
\newblock Refined commonsense knowledge from large-scale web contents.
\newblock \emph{arXiv preprint arXiv:2112.04596}.

\bibitem[{Petroni et~al.(2019)Petroni, Rockt{\"a}schel, Lewis, Bakhtin, Wu,
  Miller, and Riedel}]{Petroni2019LanguageMA}
Fabio Petroni, Tim Rockt{\"a}schel, Patrick Lewis, Anton Bakhtin, Yuxiang Wu,
  Alexander~H. Miller, and Sebastian Riedel. 2019.
\newblock Language models as knowledge bases?
\newblock \emph{EMNLP}.

\bibitem[{Qin and Eisner(2021)}]{Qin2021LearningHT}
Guanghui Qin and Jas' Eisner. 2021.
\newblock Learning how to ask: Querying lms with mixtures of soft prompts.
\newblock In \emph{NAACL}.

\bibitem[{Romero et~al.(2019)Romero, Razniewski, Pal, Z.~Pan, Sakhadeo, and
  Weikum}]{romero2019commonsense}
Julien Romero, Simon Razniewski, Koninika Pal, Jeff Z.~Pan, Archit Sakhadeo,
  and Gerhard Weikum. 2019.
\newblock Commonsense properties from query logs and question answering forums.
\newblock In \emph{Proceedings of the 28th ACM International Conference on
  Information and Knowledge Management}, pages 1411--1420.

\bibitem[{Santos et~al.(2022)Santos, Cola{\c{c}}o, Nielsen, Niu, Strauss,
  Geyer, Coscia, Albrechtsen, Mundt, Jensen et~al.}]{santos2022knowledge}
Alberto Santos, Ana~R Cola{\c{c}}o, Annelaura~B Nielsen, Lili Niu, Maximilian
  Strauss, Philipp~E Geyer, Fabian Coscia, Nicolai J~Wewer Albrechtsen, Filip
  Mundt, Lars~Juhl Jensen, et~al. 2022.
\newblock A knowledge graph to interpret clinical proteomics data.
\newblock \emph{Nature Biotechnology}, 40(5):692--702.

\bibitem[{Sap et~al.(2019)Sap, Bras, Allaway, Bhagavatula, Lourie, Rashkin,
  Roof, Smith, and Choi}]{Sap2019ATOMICAA}
Maarten Sap, Ronan~Le Bras, Emily Allaway, Chandra Bhagavatula, Nicholas
  Lourie, Hannah Rashkin, Brendan Roof, Noah~A. Smith, and Yejin Choi. 2019.
\newblock Atomic: An atlas of machine commonsense for if-then reasoning.
\newblock \emph{ArXiv}, abs/1811.00146.

\bibitem[{Shi et~al.(2019)Shi, Chen, Zhang, and Zhang}]{Shi2019NeuralLN}
Shaoyun Shi, Hanxiong Chen, Min Zhang, and Yongfeng Zhang. 2019.
\newblock Neural logic networks.
\newblock \emph{ArXiv}, abs/1910.08629.

\bibitem[{Shin et~al.(2020)Shin, Razeghi, IV, Wallace, and
  Singh}]{Shin2020ElicitingKF}
Taylor Shin, Yasaman Razeghi, Robert L~Logan IV, Eric Wallace, and Sameer
  Singh. 2020.
\newblock Eliciting knowledge from language models using automatically
  generated prompts.
\newblock \emph{EMNLP}.

\bibitem[{Speer et~al.(2017)Speer, Chin, and Havasi}]{Speer2017ConceptNet5A}
Robyn Speer, Joshua Chin, and Catherine Havasi. 2017.
\newblock Conceptnet 5.5: An open multilingual graph of general knowledge.
\newblock In \emph{AAAI}.

\bibitem[{Stich(1979)}]{stich1979animals}
Stephen~P Stich. 1979.
\newblock Do animals have beliefs?
\newblock \emph{Australasian Journal of Philosophy}, 57(1):15--28.

\bibitem[{Tan et~al.(2020)Tan, Qin, Xing, and Hu}]{tan2020summarizing}
Bowen Tan, Lianhui Qin, Eric Xing, and Zhiting Hu. 2020.
\newblock Summarizing text on any aspects: A knowledge-informed
  weakly-supervised approach.
\newblock In \emph{Proceedings of the 2020 Conference on Empirical Methods in
  Natural Language Processing (EMNLP)}, pages 6301--6309.

\bibitem[{Tandon et~al.(2014)Tandon, De~Melo, Suchanek, and
  Weikum}]{tandon2014webchild}
Niket Tandon, Gerard De~Melo, Fabian Suchanek, and Gerhard Weikum. 2014.
\newblock Webchild: Harvesting and organizing commonsense knowledge from the
  web.
\newblock In \emph{Proceedings of the 7th ACM international conference on Web
  search and data mining}, pages 523--532.

\bibitem[{Wang et~al.(2021{\natexlab{a}})Wang, Shen, Long, Zhou, and
  Chang}]{Wang2021StructureAugmentedTR}
Bo~Wang, Tao Shen, Guodong Long, Tianyi Zhou, and Yi~Chang. 2021{\natexlab{a}}.
\newblock Structure-augmented text representation learning for efficient
  knowledge graph completion.
\newblock \emph{Proceedings of the Web Conference 2021}.

\bibitem[{Wang et~al.(2019)Wang, Zhang, Zhao, Li, Xie, and
  Guo}]{Wang2019MultiTaskFL}
Hongwei Wang, Fuzheng Zhang, Miao Zhao, Wenjie Li, Xing Xie, and Minyi Guo.
  2019.
\newblock Multi-task feature learning for knowledge graph enhanced
  recommendation.
\newblock \emph{The World Wide Web Conference}.

\bibitem[{Wang et~al.(2022)Wang, Feng, Hasegawa-Johnson, and
  Yoo}]{wang-etal-2022-self}
Liming Wang, Siyuan Feng, Mark Hasegawa-Johnson, and Chang Yoo. 2022.
\newblock \href {https://doi.org/10.18653/v1/2022.acl-long.553}
  {Self-supervised semantic-driven phoneme discovery for zero-resource speech
  recognition}.
\newblock In \emph{Proceedings of the 60th Annual Meeting of the Association
  for Computational Linguistics (Volume 1: Long Papers)}, pages 8027--8047,
  Dublin, Ireland. Association for Computational Linguistics.

\bibitem[{Wang et~al.(2021{\natexlab{b}})Wang, Li, Wang, Parulian, Han, Ma, Tu,
  Lin, Zhang, Liu, Chauhan, Guan, Li, Li, Song, Ji, Han, Chang, Pustejovsky,
  Liem, Elsayed, Palmer, Rah, Schneider, and Onyshkevych}]{Wang2021COVID19LK}
Qingyun Wang, Manling Li, Xuan Wang, Nikolaus~Nova Parulian, Guangxing Han,
  Jiawei Ma, Jingxuan Tu, Ying Lin, H.~Zhang, Weili Liu, Aabhas Chauhan,
  Yingjun Guan, Bangzheng Li, Ruisong Li, Xiangchen Song, Heng Ji, Jiawei Han,
  Shih-Fu Chang, James Pustejovsky, David Liem, Ahmed Elsayed, Martha Palmer,
  Jasmine Rah, Cynthia Schneider, and Boyan~A. Onyshkevych. 2021{\natexlab{b}}.
\newblock Covid-19 literature knowledge graph construction and drug repurposing
  report generation.
\newblock In \emph{NAACL}.

\bibitem[{West et~al.(2022)West, Bhagavatula, Hessel, Hwang, Jiang, Le~Bras,
  Lu, Welleck, and Choi}]{west-etal-2022-symbolic}
Peter West, Chandra Bhagavatula, Jack Hessel, Jena Hwang, Liwei Jiang, Ronan
  Le~Bras, Ximing Lu, Sean Welleck, and Yejin Choi. 2022.
\newblock \href {https://doi.org/10.18653/v1/2022.naacl-main.341} {Symbolic
  knowledge distillation: from general language models to commonsense models}.
\newblock In \emph{Proceedings of the 2022 Conference of the North American
  Chapter of the Association for Computational Linguistics: Human Language
  Technologies}, pages 4602--4625, Seattle, United States. Association for
  Computational Linguistics.

\bibitem[{West et~al.(2021)West, Bhagavatula, Hessel, Hwang, Jiang, Bras, Lu,
  Welleck, and Choi}]{west2021symbolic}
Peter West, Chandra Bhagavatula, Jack Hessel, Jena~D Hwang, Liwei Jiang,
  Ronan~Le Bras, Ximing Lu, Sean Welleck, and Yejin Choi. 2021.
\newblock Symbolic knowledge distillation: from general language models to
  commonsense models.
\newblock \emph{arXiv preprint arXiv:2110.07178}.

\bibitem[{Xiong et~al.(2017)Xiong, Power, and Callan}]{Xiong2017ExplicitSR}
Chenyan Xiong, Russell Power, and Jamie Callan. 2017.
\newblock Explicit semantic ranking for academic search via knowledge graph
  embedding.
\newblock \emph{Proceedings of the 26th International Conference on World Wide
  Web}.

\bibitem[{Yang et~al.(2021)Yang, Cao, Zhao, Zeng, Zhang, and
  Luo}]{yang2021constructing}
Yunrong Yang, Zhidong Cao, Pengfei Zhao, Dajun~Daniel Zeng, Qingpeng Zhang, and
  Yin Luo. 2021.
\newblock Constructing public health evidence knowledge graph for
  decision-making support from {COVID}-19 literature of modelling study.
\newblock \emph{Journal of Safety Science and Resilience}, 2(3):146--156.

\bibitem[{Yao et~al.(2019)Yao, Mao, and Luo}]{Yao2019KGBERTBF}
Liang Yao, Chengsheng Mao, and Yuan Luo. 2019.
\newblock Kg-bert: Bert for knowledge graph completion.
\newblock \emph{ArXiv}, abs/1909.03193.

\bibitem[{Zhang et~al.(2020{\natexlab{a}})Zhang, Khashabi, Song, and
  Roth}]{Zhang2020TransOMCSFL}
Hongming Zhang, Daniel Khashabi, Yangqiu Song, and Dan Roth.
  2020{\natexlab{a}}.
\newblock Transomcs: From linguistic graphs to commonsense knowledge.
\newblock In \emph{IJCAI}.

\bibitem[{Zhang et~al.(2020{\natexlab{b}})Zhang, Liu, Pan, Song, and
  Leung}]{zhang2020aser}
Hongming Zhang, Xin Liu, Haojie Pan, Yangqiu Song, and Cane Wing-Ki Leung.
  2020{\natexlab{b}}.
\newblock Aser: A large-scale eventuality knowledge graph.
\newblock In \emph{Proceedings of the web conference 2020}, pages 201--211.

\bibitem[{Zhong and Chen(2020)}]{zhong2020frustratingly}
Zexuan Zhong and Danqi Chen. 2020.
\newblock A frustratingly easy approach for entity and relation extraction.
\newblock \emph{arXiv preprint arXiv:2010.12812}.

\bibitem[{Zhong et~al.(2021)Zhong, Friedman, and Chen}]{Zhong2021FactualP}
Zexuan Zhong, Dan Friedman, and Danqi Chen. 2021.
\newblock Factual probing is [mask]: Learning vs. learning to recall.
\newblock \emph{NAACL}.

\bibitem[{Zhu et~al.(2020)Zhu, Rawat, Zaheer, Bhojanapalli, Li, Yu, and
  Kumar}]{Zhu2020ModifyingMI}
Chen Zhu, Ankit~Singh Rawat, Manzil Zaheer, Srinadh Bhojanapalli, Daliang Li,
  Felix~X. Yu, and Sanjiv Kumar. 2020.
\newblock Modifying memories in transformer models.
\newblock \emph{ArXiv}, abs/2012.00363.

\end{thebibliography}
\bibliographystyle{acl_natbib}

\clearpage
\newpage
\appendix

\section{Detailed Results of Harvested Knowledge}
In Table~\ref{tab:human_setting} and Table~\ref{tab:human_model}, we show the human-annotated results of harvested knowledge in different settings. Here we list the detailed results per relation in Table~\ref{tab:results}.
\begin{table*}[h]

\caption{Detailed result of human evaluation. The numbers indicate the portions of accepted and rejected tuples. Ro-l, DB, B-b, B-l, Ro-b are short for Roberta-large, DistilBert, Bert-large, Bert-base, Roberta-base. Human, Auto, Top-1, and Multi stand for methods that use Human Prompt, Autoprompt, Top-1 Prompt (Ours), and Multi Prompts (Ours). }
\label{tab:results}
\begin{center}
\resizebox{\textwidth}{!} {
\begin{tabular}{@{}r c c c c c c c c c c c c c c c c}
\toprule
Model                         & Ro-l       & Ro-l       & Ro-l       & Ro-l       &DB          & B-b       & B-l  & Ro-b  \\ \midrule
Prompt                        & Human       &Auto        & Top-1       &Multi       &Multi       &Multi      &Multi &Multi \\ \midrule
\textsc{business}              & 0.60/0.32  & 0.76/0.13  & 0.75/0.16  & 0.88/0.07  & 0.54/0.27  & 0.64/0.23  & 0.76/0.13  & 0.74/0.19  \\
\textsc{help}                  & 0.77/0.12  & 0.52/0.34  & 0.92/0.03  & 0.87/0.05  & 0.91/0.04  & 0.81/0.04  & 0.88/0.06  & 0.88/0.06  \\
\textsc{ingredient for}        & 0.59/0.33  & 0.33/0.59  & 0.73/0.20  & 0.71/0.24  & 0.70/0.26  & 0.55/0.40  & 0.72/0.23  & 0.51/0.40 \\
\textsc{place for}             & 0.76/0.10  & 0.41/0.36  & 0.63/0.32  & 0.89/0.07  & 0.84/0.14  & 0.78/0.18  & 0.87/0.11  & 0.88/0.09 \\
\textsc{prevent}               & 0.42/0.42  & 0.18/0.67  & 0.60/0.25  & 0.40/0.45  & 0.60/0.32  & 0.44/0.39  & 0.62/0.25  & 0.68/0.25 \\ 
\textsc{source of}             & 0.76/0.17  & 0.21/0.67  & 0.52/0.44  & 0.60/0.33  & 0.63/0.36  & 0.65/0.32  & 0.75/0.24  & 0.55/0.37 \\ 
\textsc{separated by the ocean}& 0.48/0.38  & 0.16/0.48  & 0.56/0.35  & 0.55/0.40  & 0.51/0.24  & 0.57/0.26  & 0.44/0.46  & 0.44/0.49 \\ 
\textsc{antonym}               & 0.50/0.41  & 0.10/0.83  & 0.50/0.48  & 0.55/0.44  & 0.38/0.56  & 0.41/0.56  & 0.52/0.42  & 0.75/0.22 \\ 
\textsc{featured thing}        & 0.85/0.12  & 0.38/0.40  & 0.88/0.06  & 0.89/0.10  & 0.37/0.44  & 0.44/0.40  & 0.46/0.44  & 0.65/0.20 \\ 
\textsc{need A to do B}        & 0.71/0.18  & 0.62/0.21  & 0.66/0.22  & 0.79/0.10  & 0.83/0.12  & 0.62/0.25  & 0.65/0.18  & 0.72/0.17 \\ 
\textsc{can but not good at}   & 0.52/0.34  & 0.29/0.42  & 0.61/0.19  & 0.44/0.21  & 0.51/0.31  & 0.60/0.21  & 0.64/0.22  & 0.39/0.35 \\ 
\textsc{worth celebrating}     & 0.47/0.29  & 0.23/0.51  & 0.81/0.05  & 0.85/0.08  & 0.79/0.12  & 0.74/0.14  & 0.84/0.10  & 0.83/0.10 \\ 
\textsc{potential risk}        & 0.40/0.23  & 0.31/0.45  & 0.70/0.21  & 0.76/0.19  & 0.87/0.05  & 0.66/0.22  & 0.72/0.16  & 0.79/0.08 \\ 
\textsc{A do B at}             & 0.56/0.33  & 0.14/0.55  & 0.79/0.14  & 0.97/0.03  & 0.93/0.07  & 0.93/0.05  & 0.94/0.06  & 0.94/0.06 \\ 
\textsc{Average}               & 0.60/0.27  & 0.33/0.47  & 0.69/0.22  & 0.73/0.20  & 0.67/0.24  & 0.63/0.26  & 0.70/0.22  & 0.70/0.22 \\ 

\bottomrule
\end{tabular}
}
\end{center}
\end{table*}

\section{Preprocessing of ConceptNet}
\label{sec:preprocess}
We filter out some linguistic relations (e.g. \texttt{etymologically derived from}) and some trivial relations (e.g. \texttt{related to}). We only consider the tuples with confidence higher than 1, and filter out relations comprising less than 1000 eligible tuples. We don't directly take the test set from \cite{li-etal-2016-commonsense} because they reserve a lot of tuples for training, resulting in a small and unbalanced test set.



\section{Efficient knowledge tuple search}
\label{sec:algorithm}
In the candidate entity pairs proposal step, we use the minimum token log-likelihoods (shorted
as MTL) instead of the full Equation~\ref{eqn:consistent}, which allows us to apply a pruning strategy. The pseudo-code is shown in Algorithm~\ref{alg:prune}. For simplicity of the pseudo-code, we only include the case where each entity is composed of a single token. Appendix~\ref{sec:multitoken} illustrates the processing of multi-token entities. It's worth noting that our algorithm is an exact search algorithm instead of approximated algorithms like beam search, which prevents the results from biasing towards more probable head entities.

As a running example, when we are searching for 100 entity tuples, we maintain a minimum heap to keep track of the MTL of the entity tuples. The maximum size of this heap is 100, and the  heap top can be used as a threshold for future search because it’s the 100-th largest MTL: When we are searching for a
new entity tuple, once we find the log-likelihood at any time step is lower than the threshold, we can
prune the continuous searching immediately, because this means the MTL of this tuple will never
surpass any existing tuples in the heap. If a new entity tuple is searched out without being pruned, we
will pop the heap and push the MTL of the new tuple. Intuitively, the pruning process makes sure
that the generated part of the tuple in searching is reasonable for the given prompt.
\begin{algorithm}
\small
\begin{algorithmic}
\caption{Efficient Entity Tuple Search}
\label{alg:prune}
\renewcommand{\algorithmicrequire}{\textbf{Input:}}
\renewcommand{\algorithmicensure}{\textbf{Output:}}
\Require{LM: A language model; $n_r$: The entity number for a tuple of relation $r$; $N$: maximum number of candidate tuples; $P_r$: The set of prompts describing relation $r$}
\Ensure{tuple\_list: A list of $N$ entity tuples}
\State heap $\leftarrow$ MinHeap()
\Function{DFS}{cur\_tuple, cur\_MTL}
  \State idx$\leftarrow$Count(cur\_tuple)
  \If{idx = $n_r$}
    \State heap.push((cur\_tuple, cur\_MTL))
    \If{len(heap) > N}
    \State heap.pop()
    \EndIf
\EndIf
  \For{v $\in$ Vocab(LM)}
  \State cur\_L $\leftarrow \log p_{LM}(v|cur\_tuple, P_r)$ 
  \State cur\_MTL = min(cur\_L, cur\_MTL)
  \If{Count(cur\_tuple > 0) and cur\_MTL < heap.top()}
    \Return \Comment{Pruning}
  \EndIf
  \State cur\_tuple.append(v)
  \State DFS(cur\_tuple, cur\_MTL)
  \EndFor
\EndFunction
\State DFS(EmptyList(), 0)
\State tuple\_list $\leftarrow$ list(heap)

\end{algorithmic}
\end{algorithm}

\begin{figure*}[h!]
    \centering
    \includegraphics[width=0.8\textwidth]{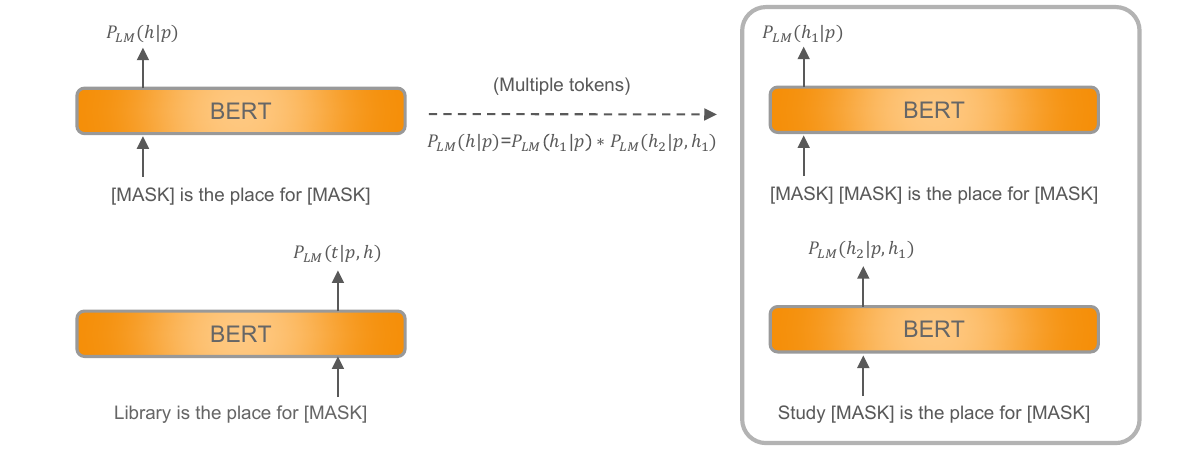}
    \caption{We demonstrate the calculation with an example where $p=$"\textsc{A is the place for B}". The left two figures shows how we calculate $P_{LM}(h|p)$ and $P_{LM}(t|p, h)$. In this example, $h=$"library" when we set both head and tail entities to have one single token. The right block shows how we calculate the conditional probability of multiple-token entities by decomposing it into two steps. In this example, the first token of the head entity $h_1=$"study".}
    \label{fig:demonstration}
\end{figure*}

\section{Detailed Experiment setting}
We use GPT-3 with the instruction "paraphrase:{sentence}" with a few examples as the off-the-shelf paraphraser. In entity pair searching, we restrict every entity to appear no more than 10 times to improve the diversity of generated knowledge and search out at most 50,000 entity tuples for each relation. We finally use various score thresholds to get the outcome KGs in different scales, including (1) 50\%: taking half of all searched-out entity pairs with higher consistency for each relation (2) base-$k$: Naturally, there are different numbers of valid tuples for different relations (e.g. tuples of  \tuple{
\ldots, capital\_of, \ldots} should not exceed 200 as that is the number of all the countries in the world). We design a relation-specific thresholding method, that is to set 10\% of the k-th consistency as the threshold (i.e., 0.1 $\times$ consistency$_k$), and retain all tuples with consistency above the threshold. We name the settings base-10 and base-100 when k is 10 and 100, respectively.
We list the truncation method applied to each variant of \textsc{RoBERTaNet} listed in Table~\ref{tab:stats}:
\begin{itemize}
    \item RobertaNet (122.2k) - Auto: base-10
    \item RobertaNet (6.7K) - ConceptNet: base-10
    \item RobertaNet (24.3K) ConceptNet: base-100
    \item RobertaNet (230K) ConceptNet: 50\%
    \item RobertaNet (2.2K) Human: base-10
    \item RobertaNet (7.3K) Human: base-100
    \item RobertaNet (23.6k) Human: 50\%
\end{itemize}
\label{sec:}

\section{Human evaluation}
\label{sec:crowdsouring}
\begin{figure*}
    \centering
    \includegraphics[width=0.95\textwidth]{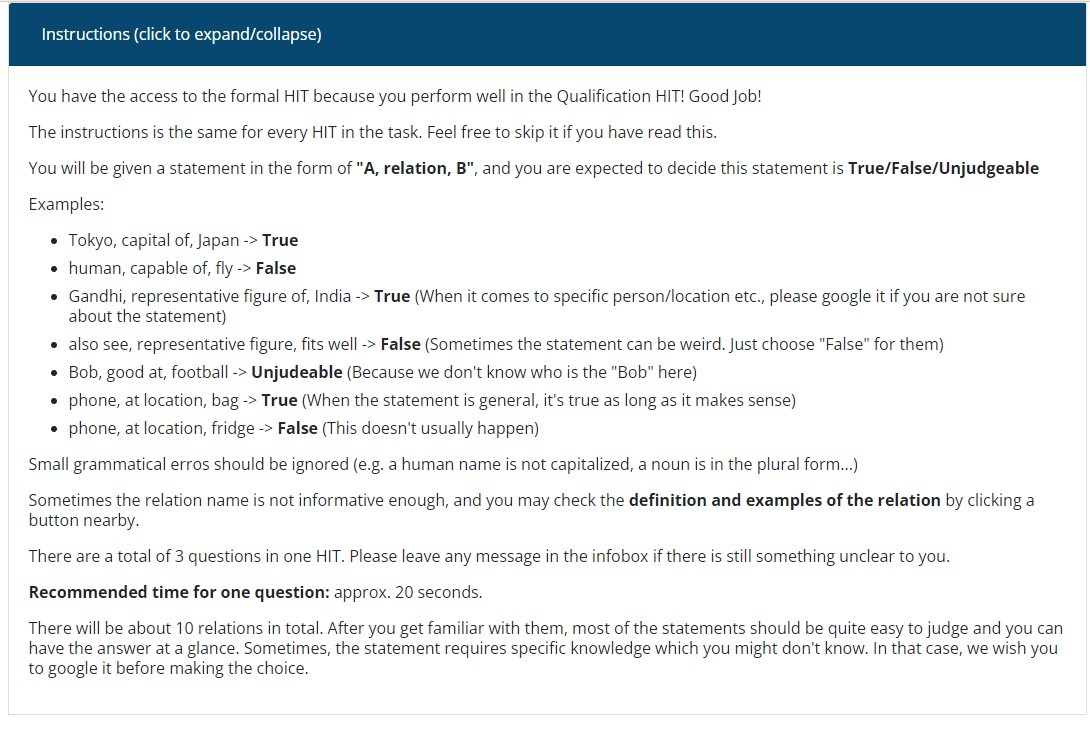}
    \caption{The instruction to annotators}
    \label{fig:inst}
\end{figure*}

\begin{figure*}
    \centering
    \includegraphics[width=0.5\textwidth]{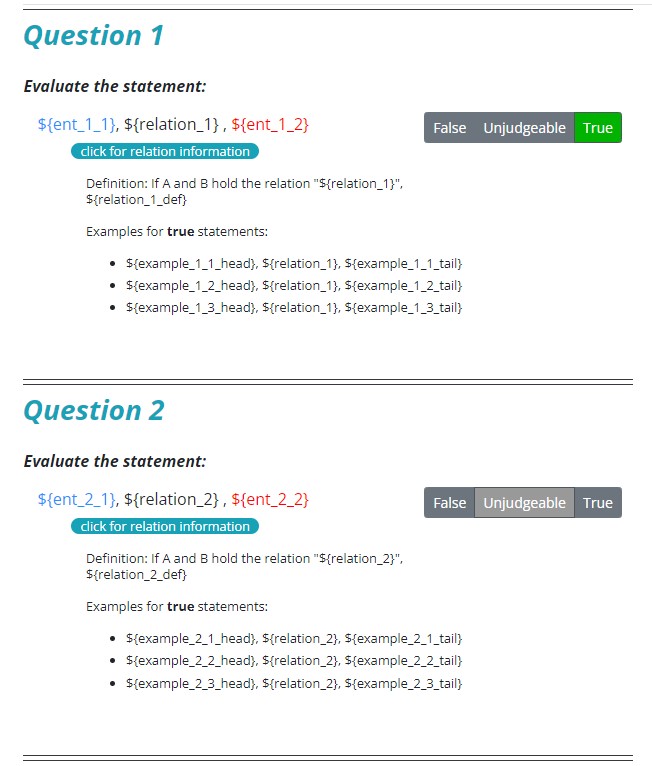}
    \caption{The questions to annotators}
    \label{fig:question}
\end{figure*}

We present the screenshot of the instruction in Figure \ref{fig:inst} and question in Figure \ref{fig:question}.
The inter-annotator agreement (Krippendorff’s Alpha) is 0.27, showing fair agreement.


\section{Compute resource}
All of our experiments are running on a single Nvidia GTX1080Ti GPU. Harvesting a knowledge graph of one relation with Roberta-large takes about one hour.

\section{The license of the assets}
All the data we used in this paper, including datasets, relation definitions, seed entity pairs, etc., are officially public resources.

\section{Potential Risks}
\label{sec:potential risks}
We identify that our system is minimal in risks. Our proposed system produce results only based on the source language models like BERT. The risks of language models are well studied and our methods do not perpetuate or add to the known risks. However, we acknowledge the methods could be applied to maliciously trained language models and discourage such uses.

\end{document}